\documentclass[runningheads,a4paper]{llncs}

\usepackage{amssymb}
\usepackage{graphicx}

\usepackage{url}
\urldef{\mailsx}\path|chaplot@cs.cmu.edu, cmaclell@cs.cmu.edu,|
\urldef{\mailsy}\path|rsalakhu@cs.cmu.edu, koedinger@cmu.edu|

\usepackage{times}
\usepackage{latexsym}

\usepackage{xcolor}
\usepackage{booktabs}
\usepackage{multirow}

\usepackage{wrapfig}

\usepackage{color}
\usepackage{amsfonts}
\usepackage{amsmath}

\newcommand{\citep}{\cite}
\newcommand{\citet}[1]{\citeauthor{#1} (\citeyear{#1})}

\newif\ifacl

\acltrue

\begin{document}

\mainmatter
\title{Learning Cognitive Models using Neural Networks}
\author{Devendra Singh Chaplot\and Christopher MacLellan\and\\ Ruslan Salakhutdinov\and Kenneth Koedinger}

\authorrunning{Learning Cognitive Models using Neural Networks}

\institute{Carnegie Mellon University,\\
5000 Forbes Avenue, Pittsburgh, PA-15217, USA\\
\mailsx\\
\mailsy\\
}

\maketitle
\begin{abstract}
A cognitive model of human learning provides information about skills a learner must acquire to perform accurately in a task domain. Cognitive models of learning are not only of scientific interest, but are also valuable in adaptive online tutoring systems. A more accurate model yields more effective tutoring through better instructional decisions. Prior methods of automated cognitive model discovery have typically focused on well-structured domains, relied on student performance data or involved substantial human knowledge engineering. In this paper, we propose Cognitive Representation Learner (CogRL), a novel framework to learn accurate cognitive models in ill-structured domains with no data and little to no human knowledge engineering. Our contribution is two-fold: firstly, we show that representations learnt using CogRL can be used for accurate automatic cognitive model discovery without using any student performance data in several ill-structured domains: Rumble Blocks, Chinese Character, and Article Selection. This is especially effective and useful in domains where an accurate human-authored cognitive model is unavailable or authoring a cognitive model is difficult. Secondly, for domains where a cognitive model is available, we show that representations learned through CogRL can be used to get accurate estimates of skill difficulty and learning rate parameters without using any student performance data. These estimates are shown to highly correlate with estimates using student performance data on an Article Selection dataset.
\end{abstract}

\section{Introduction}
A cognitive model is a computational model of human learning and problem-solving which can be used to predict human behavior and performance on the modeled problems. In this paper, we consider cognitive models in the context of intelligent tutoring systems. A cognitive model that matches student behavior provides useful information about skill difficulties and learning rates. This information often leads to better instructional decisions by aiding in problem design, problem selection and curriculum sequencing, which in turn results in more effective tutoring. A common method for representing a cognitive model is a set of \textit{Knowledge Components} (KC) \cite{vanlehn2007developing}, which represent pieces of knowledge, concepts or skills that are required for solving problems.

Cognitive models are a major bottleneck in intelligent tutor authoring and performance. Traditional ways to construct cognitive models such as structured interviews, think-aloud protocols and rational analysis requires domain expertise and are often time-consuming and error-prone \cite{li2013general}. 
Furthermore, hand-authored models can be too simplistic and are usually not verified or inconsistent with data. 

Cognitive model discovery, sometimes called ``KC model discovery'' (in Educational Data Mining) or ``Q matrix discovery'' (in Psychometrics), has been attempted through a number of different methods, but the problem remains an open, important, and interesting one. Some attempts emphasize interpretability and application of the resulting models 
\cite{cen2006learning,koedinger2013using}, while others have emphasized methods that minimize upfront human effort in feature engineering \cite{gonzalez2012dynamic,matsuda2015machine,piech2015deep,li2010machine,wang2017learning}. 

So far limited attention has focused on more ill-defined domains that are highly visual (e.g., classifying visual inputs) such as learning Chinese characters or non-discrete (probabilistic and/or with lots of exceptions) such as learning English grammar. In this paper, we tackle these ill-defined domains where complex prior perceptual skills and large amounts of background knowledge are required and where the input from the tutor is largely unstructured. We hypothesize that representations learned by machine learning techniques, which capture high-level features, can be used to create a cognitive model of human learning. We propose a novel architecture called \textit{Cognitive Representation Learner} (CogRL) to automatically extract the set of KCs required for each problem in these domains using representation learning (i.e. transforming the raw data input to a representation that can be effectively exploited in machine learning tasks). CogRL does not require any student performance data and works directly on the unprocessed problem content in the tutor. Our contribution in this paper is two-fold: firstly, we show that CogRL architecture is able to identify accurate cognitive models which outperform the baselines in a wide variety of challenging domains such as RumbleBlocks, Chinese Character, and Article Selection. This is particularly useful in domains where a good human-authored cognitive model is unavailable or difficult to construct. Secondly, for domains where a cognitive model is available, we show that learned representations can be used to get accurate estimates of skill difficulty and learning rate parameters without using any student performance data.

\section{Datasets}
\label{sec:dataset}

\subsection{RumbleBlocks}
Prior methods of cognitive model discovery only handle domains with textual problem content. We use data collected from the educational game \textit{RumbleBlocks} \cite{christel2012rumbleblocks} to test our system's ability to operate in a domain based on visual inputs. This game tasks students with building tower structures out of blocks in order to teach them basic structural stability and balance concepts. For the purposes of this study, we were less interested in modeling construction skills and more interested in modeling skills for recognizing when towers are more stable. To this end, we used a data set collected for a simplified task \cite{trestle:2016a}, where students were shown images of \textit{RumbleBlocks} towers in a randomized order and are asked to classify each tower as either ``concept 1'' or ``concept 2''. After each classification, students were provided with correctness feedback. The labels (concept 1 and concept 2) were intentionally kept vague so that students would be unable to use their prior stability and balance knowledge. The data set consists of twenty students classifying thirty towers.

\subsection{Chinese Characters}
For this domain, we use the Chinese vocabulary dataset\footnote{\small https://pslcdatashop.web.cmu.edu/DatasetInfo?datasetId=213} from the LearnLab Datashop \cite{koedinger2010data}. The problems in the dataset contain 1105 unique Chinese characters with two types of responses, English and pinyin. Pinyin refers to the English character orthography for the Chinese character pronunciation. The dataset consists of 94 students and a total of 61,323 student-item transactions. We extract the set of Chinese characters in the dataset and convert them to 16x16 images for representation learning.

\subsection{Article Selection}
For this domain, we use the data from English Article Selection task in the Intelligent Writing Tutor\footnote{\small https://pslcdatashop.web.cmu.edu/DatasetInfo?datasetId=307}. In this task, each question is a fill in the blank with three options: `a', `an' and `the'. The dataset has 84 unique problems, 79 students with a total of 4,243 student-item transactions. The dataset also provides a human-authored cognitive model with 9 Knowledge Components.

\begin{figure*}[t]
\centering
\includegraphics[width=\linewidth]{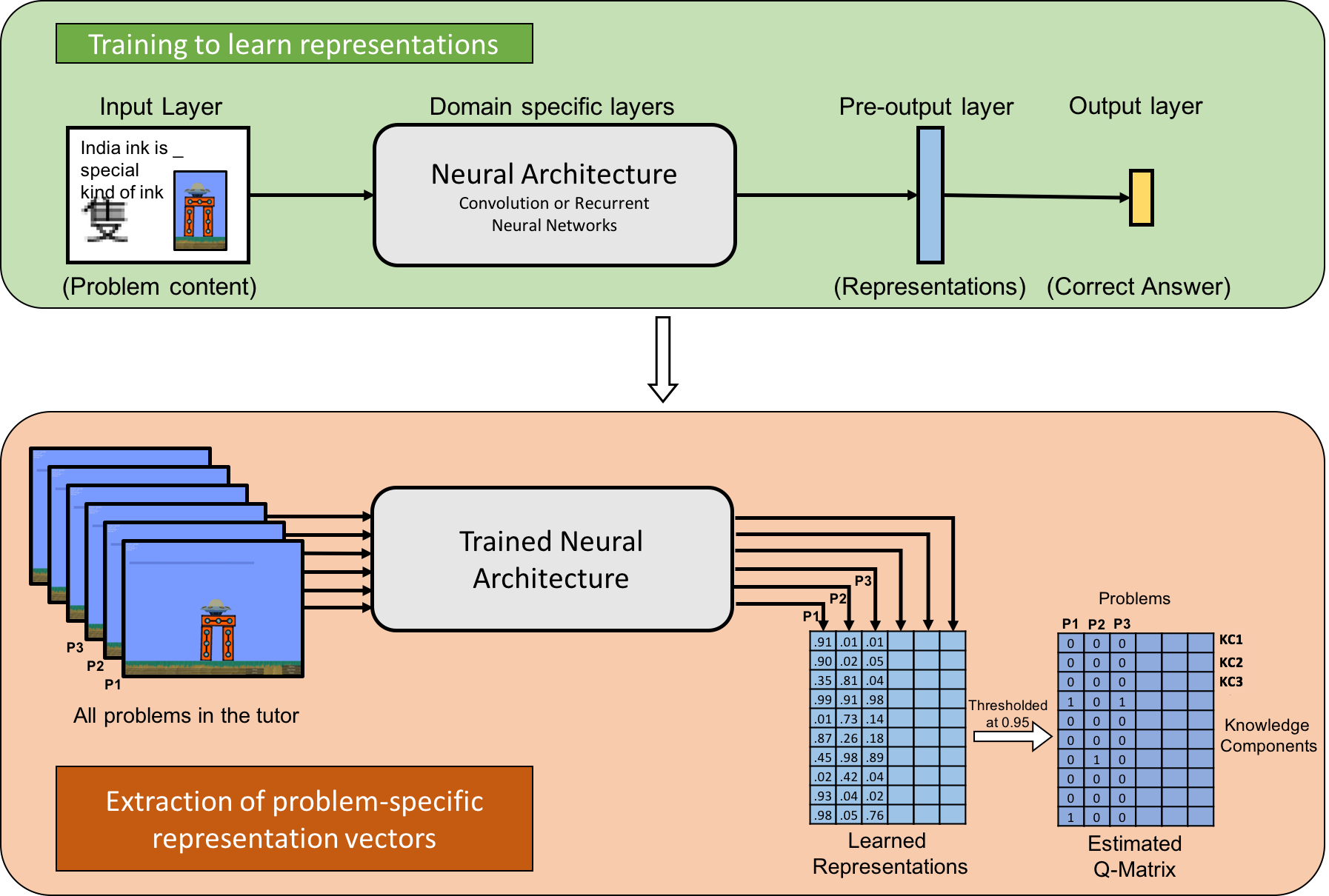}
\caption{\small Architecture of the Cognitive Representation Learner (CogRL) to estimate the Q-Matrix. For each domain, a neural architecture is chosen and trained using the problem content in the tutor and corresponding correct answer. The trained neural architecture is used to produce representation vectors in the pre-output layer for each problem. The resultant matrix is converted to a Q-Matrix by thresholding the values in the representations at 0.95.}
\label{fig:cogrl}
\end{figure*}

\section{Methods}
\label{sec:methods}
\subsection{Cognitive Representation Learner}
In this subsection, we describe the architecture of the proposed method, Cognitive Representation Learner (CogRL). For each domain, we train a Neural Architecture to predict the correct answer for the given problem in the domain. As shown in Figure~\ref{fig:cogrl}, the neural architecture, which is domain-specific, is connected to a fixed size pre-output layer, which will serve as the representations for corresponding problems. The pre-output layer is in turn connected to the output layer which predicts the correct answer for the given input problem. After training the architecture on the problems in the tutor, we use the trained model to compute the representations vectors in the pre-output layer for each problem. These representations are thresholded at 0.95 and used as columns of the estimated Q-matrix. In other words, each dimension of the learned representation constitutes a Knowledge Component in the predicted cognitive model. This cognitive model is evaluated by fitting an Additive Factors Model \cite{cen2009generalized} using the student performance data.

In the \textit{RumbleBlocks} and Chinese character datasets, the problems content is in the form of images. We use convolutional neural networks for these two datasets as they are shown to be effective in learning effective representations from pixel-based image data \cite{lecun1989backpropagation}. For the Articles Selection dataset, the challenge is that the size of the input is variable as opposed to fixed image size in the \textit{RumbleBlocks} and Chinese character datasets. Convolutional Neural Networks can not handle variable input sizes. Recurrent models are suitable for handling variable length input by treating the input as a sequence. Particularly, we use a Long Short-Term Memory Network to learn representations for the Article Selection dataset. 

In each of these architectures, we use the pre-output layer of the network as the representations for the corresponding input problem. The size of the representation or pre-output layer is not tuned to optimize the results and kept constant at 50 across all the architectures. These architectures and their training procedure are described in detail in the following subsections.

\begin{figure*}[t]
\centering
\includegraphics[width=\linewidth]{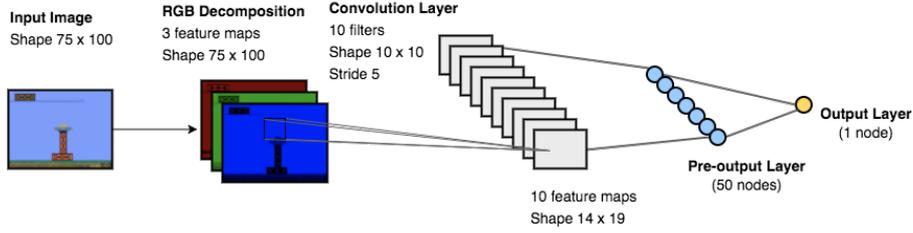}
\caption{\small Architecture of the convolutional neural network used for learning representations for the \textit{RumbleBlocks} dataset. The pre-output layer (Fully Connected Layer) is used as the learned representation for each input image.}
\label{fig:cnn}
\end{figure*}

\subsection{Convolution Neural Networks for RumbleBlocks and Chinese Characters}
Convolution Neural Networks~\cite{lecun1989backpropagation} are a type of feed-forward neural networks based on the convolution operation, which are typically used for processing visual input. Each convolutional layer consists of a set of learnable filters or kernels, which are convolved across the input image. The output is passed through a non-linear activation function, such as $\tanh$, and scaled using a learnable parameter. Each element of $y_j$ in the output of a convolution layer is calculated according to the following equation:
\begin{equation}
    y_j = g_j \tanh \Big(\sum_{i}k_{ij}*x_i\Big)
\end{equation}
where $x_i$ is the $i^{th}$ channel of input, $k_{ij}$ is the convolutional kernel, $g_j$ is a learned scaling factor and $*$ denotes the discrete convolution operation calculated using the following equation:
\begin{equation}
    (x*k)_{ij} = \sum^R_{p,q}x_{i-p,j-q}k_{p,q}
\end{equation}
where $R$ is the kernel size.

\begin{figure*}[t]
\includegraphics[width=\linewidth]{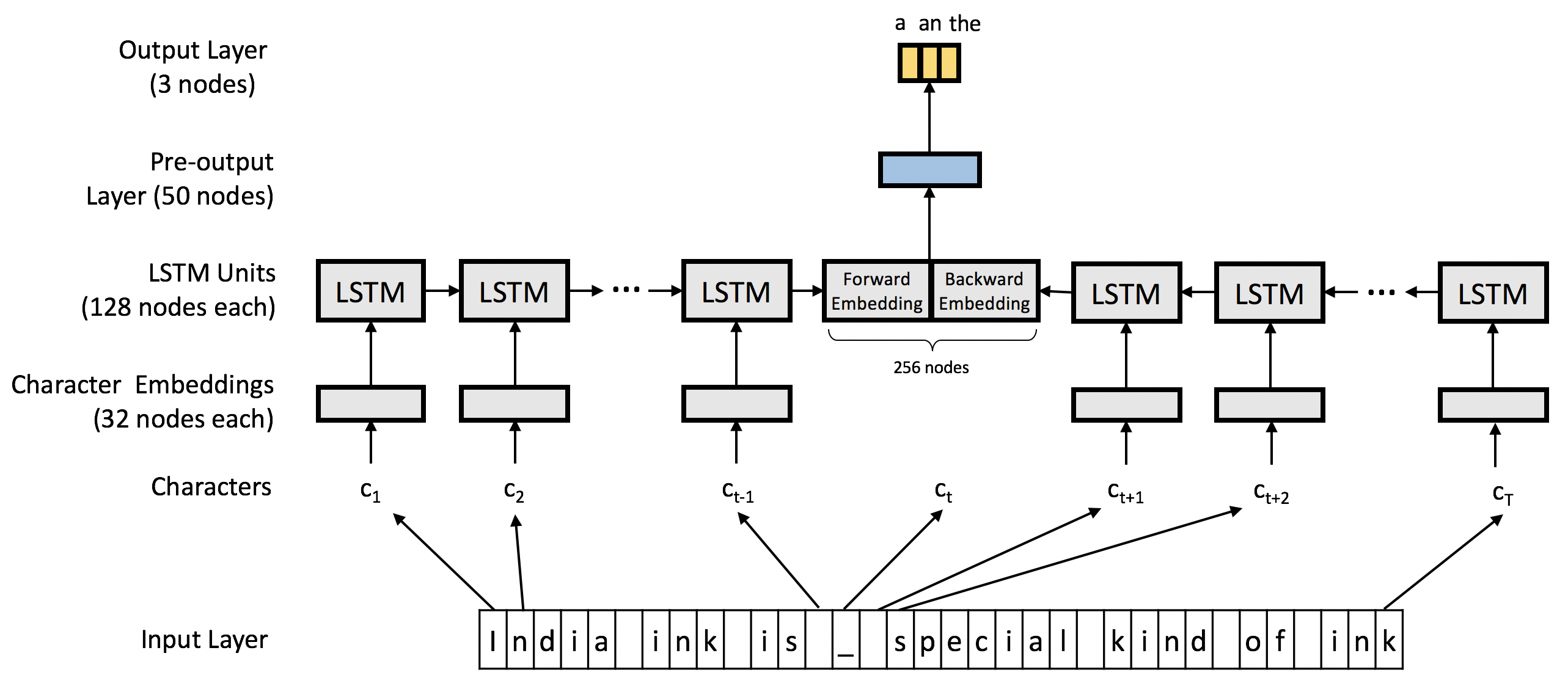}
\caption{\small Architecture of the Long Short-Term Memory (LSTM) network used for learning representations for the Article selection tutor. The pre-output layer (Fully Connected Layer) is used as the learned representation for each input sentence.}
\label{fig:lstm}
\end{figure*}

The architecture of the convolutional neural network used for learning representations for \textit{RumbleBlocks} is shown in Figure~\ref{fig:cnn}. The input image (75x100) is decomposed into red, green and blue channels which are connected to a convolutional layer consisting of 10 filters of size 10x10 with a stride of 5. The output of the convolutional layer is fully connected to a layer of 50 nodes, which in-turn is fully connected to the output layer predicting whether the configuration of rumble blocks in the input image is stable or not. The network is trained using stochastic gradient descent with a batch size of 32. After training the network, the value of the pre-output layer (50-dimensional) is used as the representation of the corresponding input image. We use the same architecture for Chinese Characters dataset except that the filter size is reduced to 4x4 with a stride 2 to match the size of a smaller 16x16 input image. 

\subsection{LSTMs for Article Selection}
Long Short-Term Memory (LSTM) \cite{hochreiter1997long} networks are a type of Recurrent Neural Networks which are suitable for sequential data with a variable size of the input sequence. In addition to the input at the current time step, nodes in a recurrent layer also receive the output of the last time step as input. This recurrence relation makes the output dependent on all the inputs in the sequence seen till the current time step. In addition to the hidden state in a vanilla recurrent unit, LSTM units have an extra memory vector and they can use explicit gating mechanisms to read from, write to, or reset the memory vector. Mathematically, at each LSTM unit, the following computations are made:\\
\begin{equation*}
\begin{split}
    i_t &= \sigma(W_{ii} x_t + b_{ii} + W_{hi} h_{(t-1)} + b_{hi}) \\
    f_t &= \sigma(W_{if} x_t + b_{if} + W_{hf} h_{(t-1)} + b_{hf}) \\
    g_t &= \tanh(W_{ig} x_t + b_{ig} + W_{hc} h_{(t-1)} + b_{hg}) \\
    o_t &= \sigma(W_{io} x_t + b_{io} + W_{ho} h_{(t-1)} + b_{ho}) \\
    c_t &= f_t * c_{(t-1)} + i_t * g_t \\
    h_t &= o_t * \tanh(c_t)
\end{split}
\end{equation*}        
where $h_t$ is the hidden state at time $t$, $c_t$ is the cell state or memory vector at time $t$, $x_t$ is the input at time $t$ and $i_t$, $f_t$, $g_t$, $o_t$ denote the input, forget, cell and out gates at time $t$, respectively.

The architecture used for learning representations for Article selection is shown in Figure~\ref{fig:lstm}. The input question is split into two parts around the blank. Each character in both the parts has a 32-dimensional embedding. The part before the blank is fed into the forward part of the LSTM sequentially, while the part after the blank is fed into the backward LSTM in the reverse order. At the end of the sequence, both LSTM parts are flattened and combined to a layer of 256 neurons. This layer is fully connected to a pre-output layer with 50 neurons. This layer will serve as the representation for the given input question, which is fully connected to the output layer. The network is trained with all the questions in the English IWT dataset described in Section~\ref{sec:dataset} using stochastic gradient descent with a batch size of 32. At the end of the training, for each input question, the pre-output layer embedding is stored as the feature representation of the question.

\section{Experiments \& Results}
\label{sec:experiments}
Each dimension of the representations learned using CogRL is considered to denote a Knowledge Component (KC). For each problem in a dataset, the representations are thresholded at 0.95, which means that if an element of the representation of a problem is greater than 0.95, then the problem is predicted to require the corresponding KC. This essentially creates a multi-KC Cognitive Model or a Q-Matrix whose rows are the thresholded representations for each problem. This automatically discovered cognitive model is evaluated by fitting the Additive Factors Model to the student performance data. We compare the CogRL cognitive model with two alternative theories of transfer of learning \cite{koedinger2016testing}.  One, the Faculty Transfer model, is based on faculty theory of transfer that suggests that the mind is like a muscle and generally improves with more experience \cite{Yaffe2009}.  The other, the Identical Transfer model, is based on the identical elements theory of transfer that suggest learning transfer occurs across nearly identical stimuli \cite{thorndike2013principles}. These models are implemented as follows:
\begin{itemize}
    \item Faculty Transfer: All the problems require a single common knowledge component.
    \item Identical Transfer: All the problems require a single unique knowledge component.
\end{itemize}
The proposed model is also compared with the best human expert cognitive model available with the tutoring system from which the data was collected. We use Item-stratified cross-validation Root Mean Square Error as the metric of comparison. The results shown in Table~\ref{tab:results1} indicate the CogRL cognitive model outperforms the baselines by a considerable margin, 0.444 vs 0.465 for Chinese Character, 0.449 vs 0.451 for Rumble Blocks and 0.399 vs 0.411 for Article Selection datasets. This indicates that the CogRL architecture is able to learn useful representations which constitute the underlying Knowledge Components for problems in various domains.

\setlength{\tabcolsep}{0.42em}
\begin{table*}[t]
\centering
\caption{Cross-validated RMSE values for fitting Additive Factors Model (AFM) using various Cognitive Models on three different datasets. The proposed model, CogRL, outperforms the baselines by a considerable margin.}
\label{tab:results1}
\begin{tabular}{lcccc}
\hline
\textbf{Dataset}  & \multicolumn{1}{l}{\textbf{Faculty Transfer}} & \textbf{Identical Transfer} & \multicolumn{1}{l}{\textbf{Best Human Model}} & \textbf{CogRL} \\ \hline
Chinese Character & 0.471                                         & 0.493                       & 0.465                                         & \textbf{0.444}                   \\
Rumble Blocks     & 0.451                                         & 0.537                       & 0.451                                         & \textbf{0.449}                   \\
Article Selection & 0.415                                         & 0.522                       & 0.411                                         & \textbf{0.399}                   \\ \hline
\end{tabular}
\end{table*}

\begin{figure}[t]
\centering
\includegraphics[width=\linewidth]{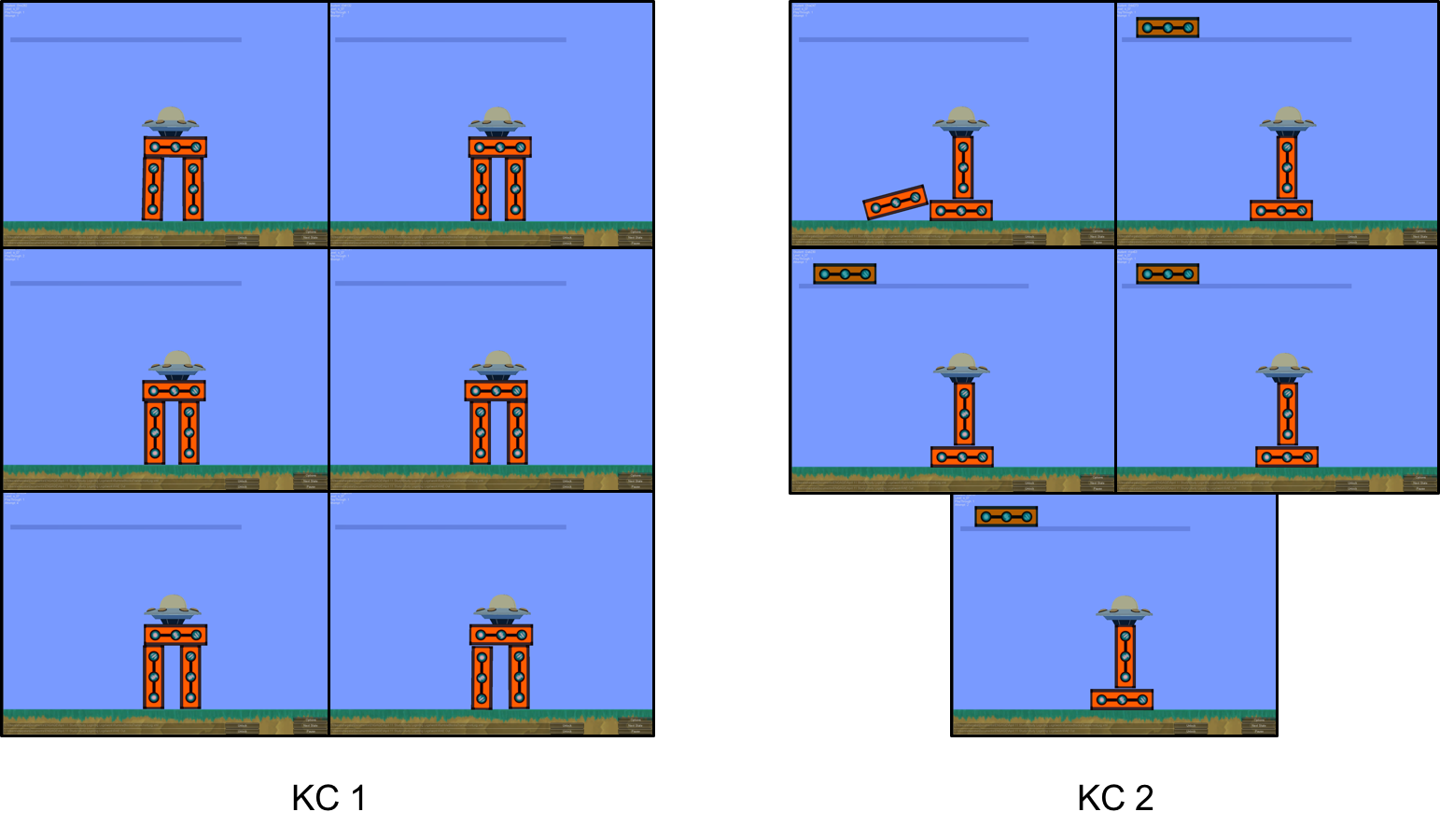}
\caption{\small Examples of non-identical \textbf{RumbleBlocks} problem images in two KCs discovered using Representation Learning. Input images which have a similar configuration of the blocks have a similar representation. Note that the images in one set are not identical, the exact position of the blocks is different in each image, although they look very similar visually.}
\label{fig:rb}
\end{figure}

We also try to analyze the representation learned by CogRL qualitatively. Figure~\ref{fig:rb} shows two sets of problem images in the Rumble Blocks domain, which require a common knowledge component. Problem images which have a similar configuration of the blocks are predicted to require a common KC. Note that the exact position of the blocks in all the shown images is very different although they look similar visually. 

\begin{figure*}[t]
\centering
\minipage{0.4\textwidth}
\includegraphics[width=\linewidth,height=\textheight,keepaspectratio]{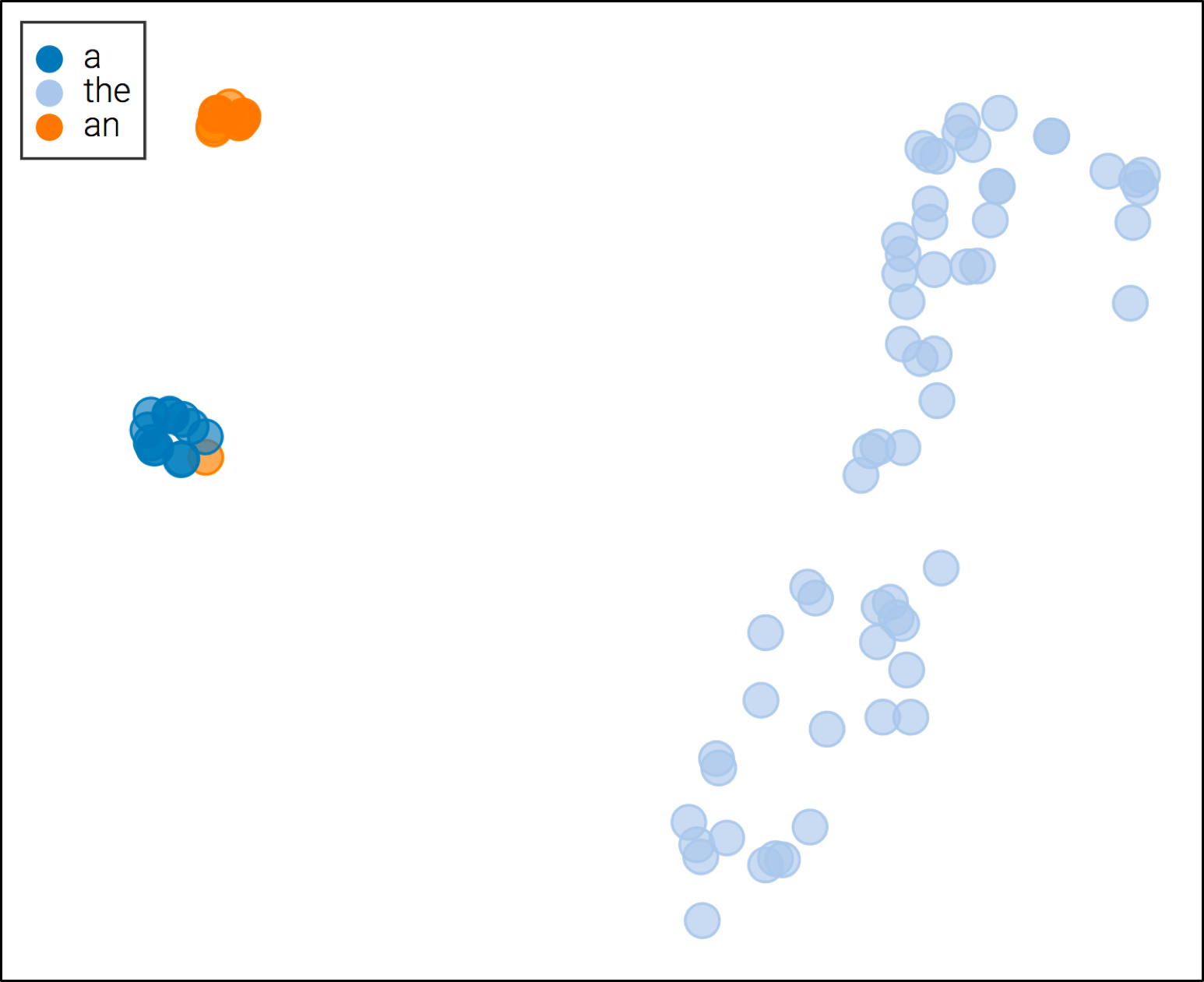}
\endminipage
\hspace{10pt}
\minipage{0.49\textwidth}
\includegraphics[width=\linewidth,height=\textheight,keepaspectratio]{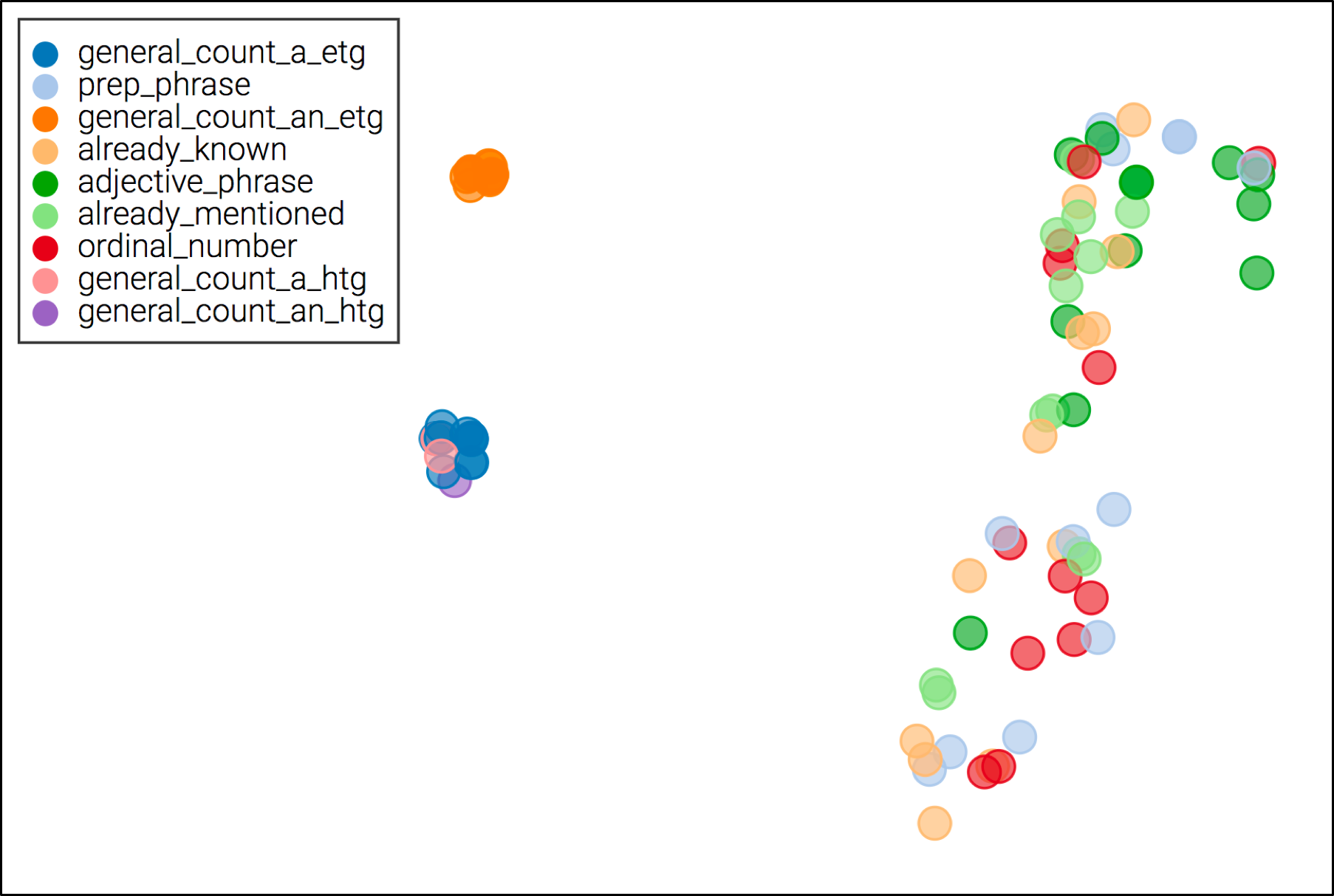}
\endminipage\hfill
\caption{t-SNE Visualization of Representations learnt for various Problems in the Article Selection dataset labelled using (a) correct answer, (b) Underlying KC in the best human-authored cognitive model.}
\label{fig:tsne}
\end{figure*}

The representations learnt for the article selection dataset are visualized using t-Distributed Stochastic Neighboring Embedding (\textit{t-SNE}) \cite{van2009learning} in Figure~\ref{fig:tsne}. t-SNE is a popular dimensionality reduction technique well suited for visualizing high dimensional data. The representations of problems are labeled according to the correct answer in Figure~\ref{fig:tsne}(a) and according to the underlying KC in the human-authored cognitive model in Figure~\ref{fig:tsne}(b). As shown in the Figure~\ref{fig:tsne}(a), the representations for problems with the same answer are very similar to each other. However, one problem with answer `an' is very similar to all problems with answer `a' rather than other problems with answer `an'. In Figure~\ref{fig:tsne}(b), we can see that this problem is the only problem in the dataset with KC ``general\_count\_an\_htg''. The problem is ``The salesman is not \_\_\_ honest man'', which belongs to this KC because the word following the blank starts with a vowel sound but not a vowel character. This makes it very similar to problems where the following word starts with a consonant and have `a' as the correct answer. Many novice learners confuse problems in this KC to have `a' as the correct answer. It is interesting to see that automatically learned representations also have this kind of relationship, which suggests that these representations might be indicative of human learning.

\section{Extension: Estimating skill difficulty and learning rates}
Intelligent Tutoring Systems are able to improve student learning across a wide range of domains by utilizing student modeling techniques (such as Additive Factors Model \cite{cen2009generalized}, Bayesian Knowledge Tracing \cite{corbett1994knowledge}, Performance Factor Analysis \cite{pavlik2009performance}) to track the skills students have acquired and to focus practice on unlearned skills. However, student modeling approaches require reasonable initial parameters in order to effectively track skill learning. In prior work, researchers have used pilot studies with fixed, non-adaptive, curriculum to empirically determine the difficulty and learning rates of skills in order to appropriately set the knowledge tracing parameters. In the previous sections, we showed that representation learning using neural architectures can be used for automatic cognitive model discovery, which is especially effective in domains where a good human-authored cognitive model is unavailable. In this section, we show that in domains where a good cognitive model is available (such as article selection), representation learning can be used to estimate the difficulty and learning rates of skills or knowledge components in the given cognitive model. For this task, we leverage the formalism of the Apprentice Learning Architecture \cite{maclellan2016apprentice} to simulate entire classroom studies for Article Selection dataset and demonstrate that empirical estimates of skill difficulty and learning rate parameters from these simulation data have high agreement with the parameters empirically estimated from human data. It is not possible to study the other domains in this context as a human-authored cognitive model is unavailable. The Apprentice learner is trained using the same sequence of problems as received by students in the Articles selection dataset. For each problem, the representations learned by CogRL are passed as input features to the Apprentice Learner. The learner fits a decision tree classifier on seen problem examples to simulate learning. The data generated using this simulation is fit using Additive Factor Model to get skill slope and intercept estimates. These estimates are compared to parameter estimates using the original student performance data.

As a baseline, we also train the Apprentice Learner using human-authored features defined by domain experts. For the article selection tutor, the domain experts defined 6 binary features, each of which is true if the following conditions hold true:
\begin{itemize}
    \item `next\_word\_starts\_with\_vowel:': Whether the word following the blank starts with a vowel. This feature approximates whether the noun following the article begins with a vowel sound.
    \item `next\_word\_ending\_st\_nd\_rd\_th': Whether the word following the blank ends with `st', `nd', `rd' or `th'. This feature is an approximation of whether the next word is an ordinal number or not.
    \item `contains\_that\_where\_who': Whether the question contains `that', `where' or `who'. This feature is an approximation of whether the noun following the article is made definite by a prepositional or an adjective phrase.
    \item `next\_word\_already\_mentioned': Whether the word following the blank is already mentioned elsewhere in the question. The feature is an approximation of whether noun that follows the article is already known or mentioned.
    \item  'next\_word\_ends\_in\_s': Whether the word after the blank ends with a `s'. This feature approximates if the noun following the article is not a singular count noun.
    \item `contains\_but\_comma': Whether the sentence contains `but' or `,'. This feature approximates whether the question has two clauses and the noun following the article is referred in the first clause and therefore, already known. For example, ``When I have watermelon, I try not to eat \_\_\_ seeds''.
\end{itemize}
Note that the article selection task is fairly complex and it is extremely difficult to author text-based features that are sufficient to answer all questions in the dataset correctly. For example, authoring text-based features which recognize vowel sound in words not starting with vowels like `honest' is extremely difficult. The features authored by the experts are shallow and are sufficient to answer only 75\% of the questions correctly.

\begin{table*}[t]
\centering
\caption{Table showing parameter estimates for the Article Selection dataset using the original student performance data, using the simulated data with Human-authored features, and using the simulated data with Representation Learning features. The parameter estimates using representation learning correlate highly with the parameter estimates using the original data.}
\label{tab:results2}
\begin{tabular}{@{}llcccccccc@{}}
\toprule
                                 &  & \multicolumn{2}{c}{\multirow{2}{*}{\begin{tabular}[c]{@{}c@{}}Original\\ data\end{tabular}}} & \multicolumn{1}{l}{} & \multicolumn{5}{c}{Apprentice Learner trained using}                                                                                                                       \\
                                 &  & \multicolumn{2}{c}{}                                                                         &                      & \multicolumn{2}{c}{\begin{tabular}[c]{@{}c@{}}Human-Authored\\ features\end{tabular}} &    & \multicolumn{2}{c}{\begin{tabular}[c]{@{}c@{}}CogRL \\ features\end{tabular}} \\ \midrule
KC Name                          &  & Intercept                                     & Slope                                        &                      & Intercept                                   & Slope                                   &    & Intercept                               & Slope                               \\ \midrule
produce\_adjective\_phrase       &  & 0.818                                         & 0.093                                        &                      & 0.881                                       & 0                                       &    & 0.563                                   & 0.156                               \\
produce\_already\_known          &  & 0.768                                         & 0.092                                        &                      & 0.547                                       & 1.208                                   &    & 0.332                                   & 0.202                               \\
produce\_already\_mentioned      &  & 0.930                                         & 0.044                                        &                      & 0.914                                       & 16.803                                  &    & 0.721                                   & 0.048                               \\
produce\_general\_count\_a\_etg  &  & 0.604                                         & 0.066                                        &                      & 0.821                                       & 0                                       &    & 0.396                                   & 0.022                               \\
produce\_general\_count\_a\_htg  &  & 0.202                                         & 0.660                                        &                      & 0.119                                       & 2.627                                   &    & 0.065                                   & 2.221                               \\
produce\_general\_count\_an\_etg &  & 0.670                                         & 0.031                                        &                      & 0.14                                        & 0.882                                   &    & 0.044                                   & 0.205                               \\
produce\_general\_count\_an\_htg &  & 0.467                                         & 0.817                                        &                      & 0                                           & 0                                       &    & 0.037                                   & 3.208                               \\
produce\_ordinal\_number         &  & 0.783                                         & 0.151                                        &                      & 0.958                                       & 0                                       &    & 0.814                                   & 0.003                               \\
produce\_prep\_phrase            &  & 0.660                                         & 0.138                                        &                      & 0.58                                        & 0.082                                   &    & 0.348                                   & 0.331                               \\ \midrule
Correlation with Original        &  & \multicolumn{1}{l}{}                          & \multicolumn{1}{l}{}                         & \multicolumn{1}{l}{} & 0.742                                       & -0.187                                  &    & 0.748                                   & 0.986                               \\ \bottomrule
\end{tabular}
\end{table*}

\subsection{Parameter estimation results}
The parameter estimates for the Article Selection dataset using 1) the original student performance data, 2) using the simulated data with Human-authored features, and 3) using the simulated data with CogRL features are shown in Table~\ref{tab:results2}. As shown in the table, the parameter estimates using CogRL features have a high correlation of 0.986 for slope, and 0.748 for intercept with the parameter estimates using the original data. This is considerably higher than the correlation of parameter estimates using Human-authored features. Most of the human-authored features are deep and result in very fast learning rate for certain KCs such as `already\_mentioned' and `already\_known', while since they don't cover all features required for learning some other KCs such as `general\_count\_an\_htg', they have a learning rate of 0. CogRL captures features essential for learning all KCs and seems to better model the struggles that learners are experiencing to acquire deep features. While it would be very difficult to train the Apprentice Learner from raw problem data, CogRL features are able to constitute the right amount of prior knowledge necessary to simulate learning in this domain. Apart from providing accurate parameter estimates without using any student performance data, the CogRL framework also minimizes the amount of human-authoring necessary to conduct simulation studies in this challenging domain. 

\section{Related Work}
\label{sec:related}
There has been a lot of interest in automating cognitive model discovery in the recent past. 
Learning Factors Analysis is a method for cognitive model evaluation and improvement which semi-automatically refines a given skill set. The improved cognitive model discovered using LFA has been used to redesign an intelligent tutoring system and shown to improve learning gains \cite{koedinger2013using}. However, LFA requires human-provided factors that require some knowledge engineering or cognitive task analysis effort. eEPIPHANY \cite{matsuda2015machine} attempts to overcome this limitation by using a collection of data-mining techniques to more automatically improve a human-crafted set of skills. LFA and ePHIPHANY both require a human-crafted set of skills as well as student performance data for cognitive model discovery and improvement. 

The requirement of student performance data makes these methods unusable for authoring a cognitive model for a new domain or a tutor with new problems. Li et al. \cite{li2010machine}  is notable prior attempt at learning a cognitive model without student performance data.  Their SimStudent learns a cognitive model by being tutored in the domain through demonstrations of correct actions and yes-no feedback on SimStudent attempts at actions. They show improved cognitive models in various domains such as algebra, stoichiometry and fraction addition \cite{li2013general}. Furthermore, the skill learning in SimStudent was also integrated with feature learning using probabilistic Context Free Grammars (pCFG) to automatically learn features to train the SimStudent \cite{li2015integrating}. However, as discussed previously, SimStudent requires structured input from the tutor interface and the learning method is mostly applicable in well-defined problem domains where minimal background knowledge is required.

Among approaches using neural networks in educational data mining, Wang et al. \cite{wang2017learning} train an LSTM to predict student's learning over time using student performance data in programming exercises. The t-SNE visualization of the hidden layer outputs of their trained neural network shows clusters of trajectories sharing some high-level properties of the programming exercise. Pardos and Dadu \cite{pardos2017imputing} use contextual representations learnt by a skip-gram model to predict missing skill from a KC model. Michalenko et al. \cite{Michalenko} use word embeddings detect misconceptions from students' responses to open-response questions.

In contrast to prior work, we tackle domains where the tutor interface is largely unstructured and problem solving requires complex prior perceptual skills and large amounts of background knowledge. Furthermore, while handling these complex domains, our methods do not require any student performance data which makes them suitable for initializing cognitive models while designing a new tutor. We also provide a method for estimating skill difficulty and learning rate without any student performance data using simulations of Apprentice learner. These estimates can be used as initialization in new tutors to provide a better estimate of mastery for each student.

\section{Conclusion \& Future Work}
\label{sec:conclusion}
We showed that representation learning using neural architectures can be used for automatic cognitive model discovery without using any student performance data, which is especially effective in domains where a good human-authored cognitive model is unavailable or authoring a good cognitive model is difficult. Qualitative analysis of representations learnt by CogRL suggests similarities with human learning. For domains where a cognitive model is available, we show that representation learning can be used to get effective estimates of skill difficulty and learning rate parameters without using any student performance data. In future, the CogRL framework can be modified to make the representations more interpretable and provide constructive feedback for improving instructional design. 

\bibliographystyle{splncs}
\bibliography{references}  

\begin{thebibliography}{10}

\bibitem{vanlehn2007developing}
VanLehn, K., Jordan, P., Litman, D.:
\newblock Developing pedagogically effective tutorial dialogue tactics:
  Experiments and a testbed.
\newblock In: Workshop on Speech and Language Technology in Education. (2007)

\bibitem{li2013general}
Li, N., Stampfer, E., Cohen, W., Koedinger, K.:
\newblock General and efficient cognitive model discovery using a simulated
  student.
\newblock In: Proceedings of the Annual Meeting of the Cognitive Science
  Society. Volume~35. (2013)

\bibitem{cen2006learning}
Cen, H., Koedinger, K., Junker, B.:
\newblock Learning factors analysis--a general method for cognitive model
  evaluation and improvement.
\newblock In: International Conference on Intelligent Tutoring Systems,
  Springer (2006)  164--175

\bibitem{koedinger2013using}
Koedinger, K.R., Stamper, J.C., McLaughlin, E.A., Nixon, T.:
\newblock Using data-driven discovery of better student models to improve
  student learning.
\newblock In: International Conference on Artificial Intelligence in Education,
  Springer (2013)  421--430

\bibitem{gonzalez2012dynamic}
Gonz{\'a}lez-Brenes, J., Mostow, J.:
\newblock Dynamic cognitive tracing: Towards unified discovery of student and
  cognitive models.
\newblock In: Proceedings of the 5th International Conference on Educational
  Data Mining. (2012)

\bibitem{matsuda2015machine}
Matsuda, N., Furukawa, T., Bier, N., Faloutsos, C.:
\newblock Machine beats experts: Automatic discovery of skill models for
  data-driven online course refinement.
\newblock International Educational Data Mining Society (2015)

\bibitem{piech2015deep}
Piech, C., Bassen, J., Huang, J., Ganguli, S., Sahami, M., Guibas, L.J.,
  Sohl-Dickstein, J.:
\newblock Deep knowledge tracing.
\newblock In: Advances in Neural Information Processing Systems. (2015)
  505--513

\bibitem{li2010machine}
Li, N., Cohen, W., Koedinger, K.R., Matsuda, N.:
\newblock A machine learning approach for automatic student model discovery.
\newblock In: Proceedings of the 4th International Conference on Educational
  Data Mining. (2011)

\bibitem{wang2017learning}
Wang, L., Sy, A., Liu, L., Piech, C.:
\newblock Learning to represent student knowledge on programming exercises
  using deep learning.
\newblock In: Proceedings of the 10th International Conference on Educational
  Data Mining. (2017)

\bibitem{christel2012rumbleblocks}
Christel, M.G., Stevens, S.M., Maher, B.S., Brice, S., Champer, M., Jayapalan,
  L., Chen, Q., Jin, J., Hausmann, D., Bastida, N.,  et~al.:
\newblock Rumbleblocks: Teaching science concepts to young children through a
  unity game.
\newblock In: Computer Games (CGAMES), 2012 17th International Conference on,
  IEEE (2012)  162--166

\bibitem{trestle:2016a}
MacLellan, C., Harpstead, E., Aleven, V., Koedinger, K.:
\newblock Trestle: A model of concept formation in structured domains.
\newblock Advances in Cognitive Systems \textbf{4} (2016)

\bibitem{koedinger2010data}
Koedinger, K.R., Baker, R.S., Cunningham, K., Skogsholm, A., Leber, B.,
  Stamper, J.:
\newblock A data repository for the edm community: The pslc datashop.
\newblock Handbook of educational data mining \textbf{43} (2010)

\bibitem{cen2009generalized}
Cen, H.:
\newblock Generalized learning factors analysis: improving cognitive models
  with machine learning.
\newblock ProQuest (2009)

\bibitem{lecun1989backpropagation}
LeCun, Y., Boser, B., Denker, J.S., Henderson, D., Howard, R.E., Hubbard, W.,
  Jackel, L.D.:
\newblock Backpropagation applied to handwritten zip code recognition.
\newblock Neural computation \textbf{1}(4) (1989)  541--551

\bibitem{hochreiter1997long}
Hochreiter, S., Schmidhuber, J.:
\newblock Long short-term memory.
\newblock Neural computation \textbf{9}(8) (1997)  1735--1780

\bibitem{koedinger2016testing}
Koedinger, K.R., Yudelson, M.V., Pavlik, P.I.:
\newblock Testing theories of transfer using error rate learning curves.
\newblock Topics in cognitive science \textbf{8}(3) (2016)  589--609

\bibitem{Yaffe2009}
Nichols, R., Yaffe, G.:
\newblock Thomas reid.
\newblock In Zalta, E.N., ed.: The Stanford Encyclopedia of Philosophy.
\newblock Winter 2016 edn. Metaphysics Research Lab, Stanford University (2016)

\bibitem{thorndike2013principles}
Thorndike, E.L.:
\newblock The principles of teaching: Based on psychology. Volume~32.
\newblock Routledge (2013)

\bibitem{van2009learning}
Maaten, L.:
\newblock Learning a parametric embedding by preserving local structure.
\newblock In van Dyk, D., Welling, M., eds.: Proceedings of the Twelth
  International Conference on Artificial Intelligence and Statistics. Volume~5
  of Proceedings of Machine Learning Research., Hilton Clearwater Beach Resort,
  Clearwater Beach, Florida USA, PMLR (16--18 Apr 2009)  384--391

\bibitem{corbett1994knowledge}
Corbett, A.T., Anderson, J.R.:
\newblock Knowledge tracing: Modeling the acquisition of procedural knowledge.
\newblock User modeling and user-adapted interaction \textbf{4}(4) (1994)
  253--278

\bibitem{pavlik2009performance}
Pavlik, P.I., Cen, H., Koedinger, K.R.:
\newblock Performance factors analysis --a new alternative to knowledge
  tracing.
\newblock In: Proceedings of the 2009 Conference on Artificial Intelligence in
  Education: Building Learning Systems That Care: From Knowledge Representation
  to Affective Modelling, Amsterdam, The Netherlands, The Netherlands, IOS
  Press (2009)  531--538

\bibitem{maclellan2016apprentice}
MacLellan, C.J., Harpstead, E., Patel, R., Koedinger, K.R.:
\newblock The apprentice learner architecture: Closing the loop between
  learning theory and educational data.
\newblock In: Proceedings of the 9th International Conference on Educational
  Data Mining. (2016)

\bibitem{li2015integrating}
Li, N., Matsuda, N., Cohen, W.W., Koedinger, K.R.:
\newblock Integrating representation learning and skill learning in a
  human-like intelligent agent.
\newblock Artificial Intelligence \textbf{219} (2015)  67--91

\bibitem{pardos2017imputing}
Pardos, Z.A., Dadu, A.:
\newblock Imputing kcs with representations of problem content and context.
\newblock In: Proceedings of the 25th Conference on User Modeling, Adaptation
  and Personalization, ACM (2017)  148--155

\bibitem{Michalenko}
Michalenko, J.J., Lan, A.S., Baraniuk, R.G.:
\newblock Data-mining textual responses to uncover misconception patterns.
\newblock stat \textbf{1050} (2017) ~30

\end{thebibliography}
\end{document}